\documentclass[11pt,english]{article}

\usepackage[latin9]{inputenc}
\usepackage{color,tikz}
\usepackage{lscape}
\usepackage{verbatim}
\usepackage{tgpagella,mathtools,float}
\usepackage{amsmath,amssymb,natbib,graphicx,url}
\usepackage[ruled]{algorithm2e}
\usepackage{hyperref}
\hypersetup{
colorlinks=true,
urlcolor=blue,
linkcolor=red,
citecolor=blue
}
\usepackage{geometry}
\newtheorem{theorem}{Theorem}
\newtheorem{lem}{Lemma}

\allowdisplaybreaks

\usepackage{times}

\begin{document}

\title{Exact and Robust Conformal Inference Methods for Predictive Machine Learning With Dependent Data}

\author{Victor Chernozhukov\thanks{email: vchern@mit.edu} \quad \quad Kaspar W\"{u}thrich\thanks{email: kwuthrich@ucsd.edu} \quad \quad Yinchu Zhu\thanks{email: yzhu6@uoregon.edu }}

\maketitle

\begin{abstract}

We extend conformal inference to general settings that allow for time series data. Our proposal is developed as a randomization method and  accounts for potential serial dependence by including  block structures in the permutation scheme such that the latter forms a group. As a result, the proposed method retains the exact, model-free validity when the data are i.i.d.\ or more generally exchangeable, similar to usual conformal inference methods.  When exchangeability fails, as is the case for common time series data, the proposed approach is approximately valid under weak assumptions on the conformity score.


\end{abstract}

\bigskip


\section{Introduction}

Suppose that we observe a times series $\{Z_t\}_{t=1}^{T_0}$, where each $Z_t=(X_t,Y_t)$ is a random variable in $\mathbb{R}^p\times \mathbb{R}$. $Y_t$ is a response variable and $X_t$ is a $p$-dimensional vector of features. We want to predict future responses $\{Y_t\}_{t=T_0+1}^{T_0+T_1}$ from future feature values $\{X_t\}_{t=T_0+1}^{T_0+T_1}$. For a pre-specified miscoverage level, we consider the problem of constructing a prediction set for  $\{Y_t\}_{t=T_0+1}^{T_0+T_1}$.

The goal and main contribution of this paper is to provide prediction sets, for which performance (coverage accuracy) bounds can be obtained in a wide range of situations, including time series data. While it is possible to design a prediction set for each problem/model, the proposed framework can be used to obtain one unified method with performance guarantees across different settings.

Our method is built on a carefully-designed randomization approach. Under the proposed methodology, we randomize the data based on a certain (algebraic) group of permutations. Note that the standard conformal prediction approach can be viewed as choosing the group to be the set of all permutations.
The key idea is to choose a group of permutations that preserve the dependence structure in the data. We do so by randomizing blocks of observations. If exchangeability does not hold, finite-sample performance bounds can still be obtained under weak conditions on the conformity score as long as transformations of the data serve as meaningful approximations for a stationary series.

Our work is closely related to the literature on randomization inference via permutations \citep{fisher1935design,rubin1984bayesianly,romano1990behavior,lehmann2005testing} and conformal inference \citep{vovk2005algorithmic,vovk2009online,lei2013distribution,vovk2013conditional,lei2014distribution,burnaev2014efficiency,balasubramanian2014conformal,lei2015conformal,lei2017distributionfree}. These papers typically exploit the i.i.d assumption to obtain the exchangeability condition under all permutations and establish model-free validity of procedures that randomize the data for general algorithms. 
The properties of these methods in the absence of exchangeability are unknown in general.  Our work makes contributions in this direction by establishing theoretical guarantees for randomization inference in the non i.i.d.\ case, covering most common types of time series models. In particular, our results cover strongly mixing processes as a special case, thereby delivering a conformal prediction method to predictive machine learning with dependent data.    The very recent work \citet{chernozhukov2017exact} explores permutations of residuals obtained from specific regression or factor models in a longitudinal data context, focusing on inference for counterfactuals in policy evaluations. By contrast, our work deals with randomization of the data and aims to robustify the conformal inference method by extending its validity to settings with dependent data. In related work, \cite{dashevskiy2008network,dashevskiy2011times} propose an interesting blocking procedure for conformal inference in times series settings, but no theoretical results are provided. By contrast, the main contribution of our work is to provide theoretical performance guarantees for conformal prediction methods when the data exhibit serial dependence.


The remainder of the paper is as follows. In Section \ref{sec: conformal prediction}, we present the setup, describe a general algorithm for constructing prediction sets, introduce the permutation schemes, and discuss two specific examples. In Section \ref{sec: theory}, we present the main theoretical properties of the proposed prediction sets. Section \ref{sec: conclusion} concludes. The appendix contains all proofs as well as a simulation experiment which demonstrates the favorable finite sample properties of the proposed approach.


\section{Conformal Inference for Dependent Data}
\label{sec: conformal prediction}

\subsection{Conformal Inference by Permutations}
Our approach is based on testing candidate values for $(Y_{T_0+1},\dots,Y_{T_0+T_1})$. Prediction sets are then constructed via test inversion. Let $y=(y_{T_0+1},\dots,y_{T_0+T_1}) $ be a hypothesized value for $(Y_{T_0+1},\dots,Y_{T_0+T_1}) $. Define the augmented data set $Z_{(y)}=\{Z_t\}_{t=1}^T$, where
\begin{equation}\label{eq: def Z}
Z_t=\begin{cases}
(Y_t,X_t) & \textrm{if}\ 1\leq t\leq T_0 \\
(y_t,X_t) & \textrm{if}\ T_0+1\leq t \leq T_0+T_1.
\end{cases}
\end{equation}

Similar to the typical conformal inference, we adopt a conformity score measure, also known as nonconformity measure \citep{vovk2009online}, which is a measurable function that maps the (augmented) data $Z_{(y)} $ to a real number. In this paper, $S(Z_{(y)} )$  denotes the conformity score and can contain  general machine learning algorithms. We shall suppress the subscript $(y)$ and write $Z$ to simplify the notation. Computing $S$ usually involves an estimator, for example, a regression estimator or a joint density. Notice that the estimators embedded in the conformity score $S$ can be either estimated  in an online manner or in the typical batch framework in statistics. Concrete examples for $S$ are provided in Section \ref{sec: examples}. 


Let $T=T_0+T_1$. Under our general setup, let $Z = \{ Z_t\}_{t=1}^T $ be arbitrary stochastic process indexed by $t \in \{1,\dots,T\}$ taking values in a sample space $\mathcal{Z}_T$. A permutation $\pi$ is bijection from $\{1,\dots,T\} $ to itself.  Let $Z^\pi = \{ Z_{\pi(t)}\}_{t=1}^T $ with $\pi \in \Pi$ be an indexed collection of arbitrary stochastic processes indexed by $t \in \{1,\dots,T\}$ taking values in  $ \mathcal{Z}_T$.  We regard these 
processes as randomized versions of $Z $. We assume that $\Pi$ includes an identity element $\mathbb{I}$ so that $Z= Z^{\mathbb{I}}$. We denote $n=|\Pi|$ and define the randomization $p$-value
$$
\hat p(y) := \frac{1}{n} \sum_{\pi \in \Pi} \mathbf{1} ( S(Z^\pi) \geq S(Z)).
$$
We also introduce the notation $\hat p =\hat{p}(y)$ with $ y=(Y_{T_0+1},\dots,Y_{T_0+T_1})$.

Given $\alpha\in (0,1)$, the predictor generates the set of $y$ with corresponding $p$-values larger than $\alpha$: 
\begin{equation}\label{eq: def S}
\mathcal{C}_{1-\alpha}=\left\{y:\  \hat p (y) > \alpha \right\}.
\end{equation}
In practice, we consider a grid of candidate values $y\in\{y^{(1)},y^{(2)},\dots,y^{(H)} \}\subset \mathbb{R}^{T_1} $; see Section \ref{sec: computation} for more discussions. We summarize this general method in Algorithm \ref{proc: generalized CP}. 

\begin{algorithm}
  \caption{Generalized Conformal Inference}\label{proc: generalized CP}
  \KwIn{Data $\{(Y_t,X_t)\}_{t=1}^{T_0} $,  $\{X_t\}_{t=T_0+1}^{T_0+T_1} $, miscoverage level $\alpha\in (0,1)$, conformity score $S(\cdot)$, permutation scheme $\Pi$}
  \KwOut{$(1-\alpha)$  confidence set   $\mathcal{C}_{1-\alpha}$}
      \For{$y\in\{y^{(1)},y^{(2)},\dots,y^{(H)} \}\subset \mathbb{R}^{T_1}  $}{
        define $Z_{(y)}$ as in \eqref{eq: def Z} \\
       compute $\hat{p}(y)$ as in \eqref{eq: def S}}
      {
        \textbf{Return} the $(1-\alpha)$  confidence set  $\mathcal{C}_{1-\alpha}=\left\{y: \hat p (y) > \alpha \right\}.$
      }
\end{algorithm}

Note that the $p$-value can also be stated in terms of order statistics. Let  $\{S^{(j)}(Z)\}_{j=1}^{n}$ denote the non-decreasing rearrangement of $\{S(Z^\pi): \pi \in \Pi\}$. Call these randomization quantiles. Observe that $$
\mathbf{1} \{\hat p  \leq \alpha\} = \mathbf{1}\{ S(Z) > S^{(k)}( Z) \},
$$
where $k = k(\alpha) =  n - \lfloor n/\alpha\rfloor = \lceil n(1-\alpha)\rceil$. 

\subsection{Designing Permutations $\Pi$ for Dependent Data}\label{sec: block permutation}

To construct valid prediction sets, we need to take into account the dependence structure in the data. We therefore design $\Pi$ to have a block structure which preserves the dependence. These blocks are allowed to be overlapping or non-overlapping. 



We start with the non-overlapping blocking scheme. Let $b$ be an integer between $T_1$ and $T$.  We split the data into $K=T/b$ blocks with each block having $b$ consecutive observations. (Here, and henceforth, we assume that the $T/b$ is an integer, for simplicity, as only very minor changes are needed when $T/b$ is not integer-valued.)  

We divide the data into $K$  non-overlapping blocks and each block contains $b$ observations.
We adopt the convention of labeling the last $b$ observations as the first block. Therefore, the $j$-th block contains observations for $t\in \{ T-jb+1,\dots,T-(j-1)b \}  $. For $1\leq j\leq K$, we define the $j$-th non-overlapping block (NOB) permutation $\pi_{j,\text{NOB}}: \{1,\dots,T\} \rightarrow \{1,\dots,T\} $ via 
\begin{equation} \label{eq: non-overlapping permutation}
t \mapsto \pi_{j,\text{NOB}}(t)= \left . \begin{cases}
t+(j-1)b & \text{ if}\ 1\le t\le T-(j-1)b\\
t+(j-1)b-T &  \text{ if}\ T-(j-1)b+1\le t \le T\\
\end{cases} \right | t = 1,\dots,T.
\end{equation}
Using the modulo operation, we can write $\text{mod}\,(t+(j-1)b-1,T)+1 $.
The collection of all permutation is given by $\Pi_{\text{NOB}}=\{\pi_{j,\text{NOB}}:\ 1\leq j\leq K\} $. Clearly, $\Pi_{\text{NOB}}$ is a group (in the algebraic sense) and contains the identity map.


We also consider an overlapping blocking scheme. We construct the permutation as a composition of elements in $\Pi_{\text{NOB}}$ and cyclic sliding operation (CSO) permutations. We first define CSO permutations. For $1\leq j\leq T$, consider permutations defined by:
 $$ t \mapsto \pi_{j,\text{CSO}}(t)= \left. \begin{cases}
t+(j-1) & \text{ if}\ 1\leq t\leq T-(j-1)\\
t+(j-1)-T & \text{ if}\ T-(j-1)+1\leq t\leq T
\end{cases}  \right | \quad t=1,\dots,T. $$

The set of cyclic sliding operations is then $\Pi_{\text{CSO}}= \{\pi_{j,\text{CSO}}:\ 1\leq j\leq T \} $. Finally, an overlapping block scheme can be represented by the Minkowski composition of the two groups:
\begin{equation} \label{eq: overlapping permutation}
\Pi_{\text{OB}}=\Pi_{\text{CSO}} \circ \Pi_{\text{NOB}} =\{\pi_{j_1,\text{CSO}} \circ \pi_{j_2,\text{NOB}}:\ 1\leq j_1 \leq T,\ 1\leq j_2 \leq K  \} .
\end{equation}
The set of overlapping block permutations, $\Pi_{\text{OB}}$, also forms a group and contains the identity map. 
Observe that $\Pi_{\text{CSO}} $ is a group and gives back $\Pi_{\text{CSO}}$ when composed with  $\Pi_{\text{NOB}} $, so that $\Pi_{\text{OB}} = \Pi_{\text{CSO}}$.

\subsection{Examples} \label{sec: examples}

\subsubsection{Penalized Regression}\label{sec: penalized regression}

Assume that the data are drawn from the model
\begin{equation}
Y_t=X_t'\beta+\varepsilon_t, \quad 1\le t \le T, \label{eq: linear model}
\end{equation}
where $\varepsilon_t$ is mean-zero stationary stochastic process and $\beta \in \mathbb{R}^p$ is a coefficient vector. We estimate $\beta$ based on the augmented dataset $Z$ using penalized regression
\[
\hat\beta (Z) = \arg\min_{\beta\in \mathbb{R}^p} \frac{1}{T}\sum_{t=1}^T\left(Y_t-X_t'\beta \right)^2+ \text{pen}(\beta),
\]
where $\text{pen}(\cdot)$ is a penalty function. Popular penalty functions include $\ell_1$-norm (LASSO), $\ell_2$-norm (ridge regression) and non-convex penalties (e.g., SCAD). 
We define  the fitted residual as
\begin{eqnarray*}
\hat{\varepsilon}_t(Z) =Y_t-X_t'\hat\beta (Z), \quad t=1,\dots,T.
\end{eqnarray*}
Consider the following residual-based conformity score, which operates on the last $T_1$ elements of the residual vector:
\begin{equation}\label{eq: def S pen reg}
S(Z)=  \left(\sum_{t=T_0+1}^{T_0+T_1} \left|\hat{\varepsilon}_{t}(Z) \right|^p\right)^{1/p}.
\end{equation}

If $T_1=1$ and $p=1$, this conformity score corresponds to the absolute value of the last residual, $S(Z)=|\hat{\varepsilon}_T(Z)|$ as in \citet{lei2017distributionfree}. A natural choice for the block size is $b=T_1$ such that $K=T/T_1$. If $\hat{\beta}(Z)$ is invariant to permutations of the data (which is the case for most regression methods), $p$-values based non-overlapping and overlapping block permutations can be computed by permuting the fitted residuals.


\subsubsection{Autoregressive Models and Neural Networks}
Assume that the data are generated by a $K$-th order linear autoregressive model
\[
Y_t=\sum_{k=1}^K\rho_k \mathrm{L}^k(Y_{t})+ \varepsilon_t, \quad t=1,...,T, \quad Y_0,\dots,Y_{-K+1} ~~\text{given.} 
\]
where $\mathrm{L}^k$ is the lag operator. We use least squares to obtain an estimate of the vector of autoregressive coefficients $\rho=\left(\rho_1,\dots,\rho_K\right)$ based on the augmented data $Z$. Denote this estimator as  $\hat{\rho}=\left(\hat\rho_1,\dots,\hat\rho_K\right)$ and define the fitted residuals as  
\[
\hat\varepsilon_t\left(Z\right)=Y_t-\sum_{k=1}^K\hat\rho_k \mathrm{L}^k(Y_{t}), \quad 1\le t\le T,
\]

More generally, we can consider nonlinear autoregressive models
\[
Y_{t}=\rho\left(Y_{t-1},\ldots,Y_{t-K}\right)+\varepsilon_{t},  \quad 1\le t\le T, \quad Y_0,\dots,Y_{-K+1} ~~\text{given}, 
\]
where $\rho $ is a nonlinear function. Such models arise when using neural networks for predictive time series modeling \citep[e.g.,][]{chen1999improved,chen2001semiparametric}. We allow $\rho $ to be parametric, nonparametric or semi-parametric. Let $\hat{\rho}$ be a suitable estimator for $\rho$, obtained based on the augmented data $Z$. Define the fitted residuals as
\[
\hat\varepsilon_t\left(Z\right)=Y_t-\hat{\rho}\left(Y_{t-1},\ldots,Y_{t-K}\right),  \quad 1\le t\le T.
\]

For both the linear and the nonlinear models, we choose a residual-based conformity score
\[
S(Z)=  \left(\sum_{t=T_0+1}^{T_0+T_1} \left|\hat{\varepsilon}_t\left(Z\right) \right|^p\right)^{1/p}.
\]
A natural choice for the block size is $b=T_1$ such that there are $K=T/T_1$ blocks in total.


\subsection{Computational Aspects}\label{sec: computation}
Here we briefly discuss the computational aspects of the proposed procedure. As we have noted in Algorithm \ref{proc: generalized CP}, the procedure is performed on a chosen grid of values for $y$. In choosing the grid, we should take into account the computational burden. For nonconformity measure obtained by estimating a model, an important factor is how many times we need to implement the learning algorithm. For a given $y$, computing $S(Z^{\pi}) $ might in general require running the estimation algorithm for each $\pi \in \Pi$; as a result, one needs to train the model $|\Pi|$ times to compute  $\hat{p}(y)$ for a given $y$. In this case, it might not be realistic to choose a large number of points ($H$ in Algorithm \ref{proc: generalized CP}). However, it is often possible to exploit the structure of the problem to reduce the computational burden. Consider for instance the penalized regression setting of Section \ref{sec: penalized regression} where the non-conformity measure is a transformation of the regression residuals. Provided that the estimator of $\beta$ is invariant under permutations of the data, we only need to implement the training algorithm once for a given $y$. The dimensionality of $y$ ($T_1$) also plays a role in determining the grid. For problems that only consider $T_1=1$, we can simply choose a equal-spaced grid on an interval. For $T_1>1$, the choice of the grid might require extra care to keep the computation feasible; from our experience, this typically involves exploiting the nature of the problem at hand and hence is a case-by-case analysis.

To further reduce the computational burden, we would like to design an algorithm that does not have to train the model for every point of $y$ in the grid. One referee brought to our attention the inductive conformal prediction approach \citep[e.g.,][]{papadopoulos2007conformal}. The idea is to separate the data into ``proper training set'' and ``calibration set'', to train the model only once on the former and to conduct the permutations only on the latter. Since we do not need to start from scratch for each candidate $y$, computing the confidence set only requires running the training algorithm once. Notice that we can view this as a special case of our general permutation framework: this amounts to restricting the set of permutations to those that only permute the indices on the ``calibration set'' and keep the indices on the ``proper training set'' unchanged. Hence, theoretical results developed in our work can be used to justify inductive conformal predictions for dependent data.

\section{Theory: General Results on Exact and Approximate Conformal Inference}
\label{sec: theory}

We now provide theoretical guarantees for the proposed method. When the data are exchangeable, the proposed approach exhibits model-free and exact finite-sample validity, similar to the existing conformal inference methods. When the data are serially dependent and exchangeability is violated, our method retains approximate finite sample validity under weak assumptions on the conformity score as long as transformations of the data serve as meaningful approximations for a stationary series.


\subsection{Exact Validity}

The key insight from the randomization inference literature is to exploit the exchangeability in the data. Since we can cast conformal inference approaches as randomizing in the set $\Pi$, we can analyze the proposed generalized conformal inference method (Algorithm \ref{proc: generalized CP}) by examining the exchangeability
and the quantile invariance property (implied by $\Pi$ being a group). 

\begin{theorem}[General Exact Validity] \label{prop: general exact validity}
Suppose that $\{ Z^\pi\}$ has an exchangeable distribution under permutations $\pi \in \Pi$. Consider any fixed  $\Pi$  such that  the randomization $\alpha$-quantiles are invariant surely, namely
$$ S^{(k(\alpha))}(Z^\pi) = S^{(k(\alpha))}(Z), \text{ for all } \pi \in \Pi.$$
The latter condition holds when $\Pi$ is a group. Or, more generally, suppose that  surely \begin{equation}\label{quant bound}
 S^{(k(\alpha))}(Z^\pi)  \geq S^{(k)}(Z),  \text{ for all } \pi \in \Pi.
 \end{equation}
Then
$$
P(\hat p  \leq \alpha) =  P(  S(Z) > S^{(k)}(Z) ) \leq \alpha \ \ \text{ 
and } \ \ P((Y_{T_0+1},\dots,Y_{T_0+T_1}) \in \mathcal{C}_{1-\alpha})\geq 1-\alpha.$$
\end{theorem}

This result follows from standard arguments for randomization inference, see \citet{romano1990behavior}. To the best of our knowledge, this is the weakest condition under which one can obtain model-free validity of conformal inference. A sufficient condition for exchangeability is that the data is i.i.d. (exchangeable) and that $\Pi$ is a group.



\subsection{Approximate Validity}
When a meaningful choice of $S$ is available, we can relax the exchangeability condition and expect to achieve certain optimality. Let $S_* $ be an oracle score function, which is typically an unknown population object. For example,  $S_*$ can be a transformation of the true population conditional distribution of $(y_{T_0+T_1},\dots,y_{T_0+T_1}) $ given $ (X_{T_0+T_1},\dots,X_{T_0+T_1}) $. In a regression setup, $S_* $ might be measuring the magnitude of the error terms; in the example of Section \ref{sec: penalized regression}, $S_*$ would be the analogous of $S$ defined in \eqref{eq: def S pen reg} with true residuals:
\begin{equation} \label{eq: example S star}
S_*(Z)=  \left(\sum_{t=T_0+1}^{T_0+T_1} \left|{\varepsilon}_{t}(Z) \right|^p\right)^{1/p},
\end{equation}
where $\varepsilon_t(Z)=Y_t-X_t'\beta$.  We show that when $S$ consistently approximates the oracle score $S_*$, the resulting confidence set is valid and approximately equivalent to inference using the oracle score.

For approximate results, assume that
the number of randomizations
becomes large, $n = |\Pi|\rightarrow\infty$ (in examples above, this is caused by $T_0 \to \infty$). Let $\{\delta_{1n}, \delta_{2n}, \gamma_{1n}, \gamma_{2n}\}$ be sequences of numbers converging to zero, and assume the following conditions.
\begin{itemize} 
\item [(E)] With
probability $1-\gamma_{1n}$:  the randomization distribution 
$$\tilde{F}(x):=\frac{1}{n}\sum_{\pi\in\Pi}\mathbf{1}\{S_*(Z^\pi)< x\},$$
is \textit{approximately ergodic} for $F(x)=P\left(S_*(Z)< x\right)$, namely 
$$
\sup_{x\in\mathbb{R}}\left|\tilde{F}(x)-F\left(x\right)\right|\leq\delta_{1n}.
$$
\end{itemize}
\begin{itemize} 
\item [(A)] With
probability $1-\gamma_{2n}$, estimation errors are small:
\begin{itemize}
\item[(1)] the mean squared error is small, $n^{-1}\sum_{\pi\in\Pi}\left[S(Z^\pi)-S_*(Z^\pi)\right]^{2}\leq\delta_{2n}^{2};$ 
\item[(2)] the pointwise error at $\pi=\mathrm{Identity}$ is small, $|S(Z)-S_*(Z)|\leq\delta_{2n}$; 
\item[(3)] The pdf of $S_*(Z)$ is bounded above by a constant $D$. 
\end{itemize}
\end{itemize}

Condition (A) states the precise requirement for the quality of approximating the oracle $S_*(Z^\pi)$ by $S(Z^\pi) $. When we view $S$ as an estimator for $S_*$,  we merely require pointwise consistency and consistency in the prediction norm. This condition can be easily verified for many estimation methods under appropriate model assumptions. For example, in  sparse high-dimensional linear models, we can invoke well-known results such as \citet{bickel2009simultaneous}. For linear autoregressive models, sufficient conditions follow from standard results in \citet{hamilton1994time} and \citet{brockwell2013time}. For neural networks, sufficient conditions can be derived from results in \citet{chen1999improved}. 

Condition (E) is an ergodicity condition, which states that permuting the oracle conformity scores provides a meaningful approximation to the unconditional distribution of the oracle conformity score. In Section \ref{subsec: primitive conditions}, we show that Condition (E) holds for strongly mixing time series using the groups
of blocking permutations defined in Section \ref{sec: block permutation}. For regression problems,  $S_* $ is typically constructed as a transformation of the regression errors.  

The next theorem shows that, under conditions (A) and (E), the proposed generalized conformal inference method is approximately valid.






\begin{theorem}[\textbf{Approximate General Validity of Conformal Inference}]\label{thm: high level new}
Under the approximate ergodicity condition (E) and the small error condition (A), the approximate conformal p-value is approximately uniformly distributed, that is, it obeys for any $\alpha\in(0,1)$ 
$$\left|P\left(\hat{p}\leq\alpha\right)- \alpha \right|  \leq 6\delta_{1n}+ 4 \delta_{2n}+ 2 D(\delta_{2n}+ 2\sqrt{\delta_{2n}}) +\gamma_{1n}+\gamma_{2n}
$$
and the conformal confidence set has approximate coverage $1-\alpha$, namely
$$ \left| P((Y_{T_0+1},\dots,Y_{T_0+T_1}) \in \mathcal{C}_{1-\alpha}) -(1-\alpha)\right| \leq  6\delta_{1n}+ 4 \delta_{2n}+ 2 D(\delta_{2n}+ 2\sqrt{\delta_{2n}}) +\gamma_{1n}+\gamma_{2n}. $$
\end{theorem}



Under further stronger conditions on the estimation quality, the generalized conformal prediction $\mathcal{C}_{1-\alpha} $ achieves an oracle property in volume. Let $ \mathcal{C}_{1-\alpha}^*=\{y: 1-F(S_*(Z)) > \alpha \} $ be the oracle prediction set. Let $\mu(\cdot)$ denote the Lebesgue measure. When such an oracle prediction set has continuity in the sense that $\mu(\{y:\ |F(S_*(y))-(1-\alpha)|\leq \varepsilon \}) \rightarrow 0$ as $\varepsilon\rightarrow 0$, we can show that shrinking errors in approximating $\{S_*(Z^\pi)\}_{\pi\in \Pi} $ by $\{S(Z^\pi)\}_{\pi\in \Pi} $ implies that $\mu(\mathcal{C}_{1-\alpha} \bigtriangleup \mathcal{C}_{1-\alpha}^*)$ decays to zero,  where $ \bigtriangleup  $ denotes the symmetric difference of two sets. Such results have been established by \cite{lei2013distribution} among others for specific models under i.i.d.\ data.

\subsection{Approximate Ergodicity for Strongly Mixing Time Series with Blocking Permutations $\Pi$}
\label{subsec: primitive conditions}

In the blocking schemes discussed in Section \ref{sec: block permutation}, we can view $\{S_*(Z^\pi) \}_{\pi\in \Pi} $ as a time series $\{u_t\} $. Recall $S_*(Z)$ defined in (\ref{eq: example S star}) for the penalized regression example in Section \ref{sec: penalized regression}. By non-overlapping block permutations defined in (\ref{eq: non-overlapping permutation}) with $b=T_1$, we can see that $\{S_*(Z^\pi) \}_{\pi\in\Pi_{\text{NOB} } } $ can be rearranged as $\{u_t\}_{t=1}^K $, where 
$$
u_t=  \left(\sum_{s=T_0+(1-t)b+1}^{T_0+(2-t)b} \left|{\varepsilon}_{s} \right|^p\right)^{1/p}
$$
and $\varepsilon_s$ is the true regression residuals in (\ref{eq: linear model}). 

The situation with overlapping permutations is more complicated. Let $b=T_1$. We observe that for $1\leq k\leq K$, $\pi_{k,\text{NOB}}=\pi_{b(k-1)+1,\text{CSO}} $; for $1\leq j_1,j_2\leq T$, $\pi_{j_1,\text{CSO}}\circ \pi_{j_2,\text{CSO}}=\pi_{j_1+j_2-1-T\mathbf{1}\{j_1+j_2>T+1\},\text{CSO}} $. Therefore, for a fixed $1\leq j\leq T$, we define the  integer $q=  \lfloor (T+1-j)/b \rfloor $  and rearrange $\{S_*(Z^{\pi_{j,\text{CSO}}\circ \pi_{k,\text{NOB}}  })\}_{k=1}^K $ as follows: 

\begin{equation} \label{eq: two segments}
\left\{\left(\sum_{s=T_0+2-j-b(k-1)}^{T+1-j-b(k-1)} \left|{\varepsilon}_{s} \right|^p\right)^{1/p} \right\}_{k=1}^{q},\ \left\{\left(\sum_{s=T+T_0+2-j-b(k-1)}^{2T+1-j-b(k-1)} \left|{\varepsilon}_{s} \right|^p\right)^{1/p} \right\}_{k=q+2}^{K} 
\end{equation}
and 
$$
\left(\sum_{1\leq s \leq T+2-j-bq \text{ or } T+T_0+2-j-bq\leq s\leq T } \left|{\varepsilon}_{s} \right|^p\right)^{1/p}.
$$

Therefore, for any fixed $1\leq j\leq T$, we can rearrange $\{S_*(Z^{\pi_{j,\text{CSO}}\circ \pi_{k,\text{NOB}}  })\}_{k=1}^K $ to be two segments of stationary process in (\ref{eq: two segments}) and one extra term.

With this setup in mind, we provide a result that gives a mild sufficient condition for the ergodicity condition (E). Our result is built upon the notation of strong mixing conditions; see e.g., \cite{bradley2007introduction,rio2017asymptotic}. In our context, we define the strong mixing coefficient $\alpha_{mixing} $ for a sequence $\{u_t\}_{t=1}^{\infty} $ by
\begin{multline*}
 \alpha_{mixing}(k)= \sup \{P(A\bigcap B)-P(A)P(B):\ A\in \sigma(\{u_t:\ t\leq s\}), \\
 B \in \sigma(\{u_t:\ t\geq s+k\}), s\in \mathbb{N} \},
\end{multline*}
where $\sigma(\cdot)$ denotes the $\sigma$-algebra generated by random variables. Our formal result is stated as follows.


\begin{lem}[Mixing implies Approximate Ergodicity]
\label{lem: primitive cond for ergodicity} We consider both overlapping blocks and non-overlapping blocks. 
\begin{enumerate}
	\item Let $\Pi=\Pi_{\text{NOB}} $  the set of non-overlapping blocks defined in \eqref{eq: non-overlapping permutation}. Suppose that there exists  $\{u_t\}_{t=1}^K $ a rearrangement of $\{S_*(Z^\pi)\}_{\pi\in\Pi}$ such that $\{u_t\}_{t=1}^K $ is stationary and strong mixing with $ \sum_{k=1}^\infty \alpha_{\text{mixing}}(k) \leq M $ for a constant $M$. Then  there exists a constant $M'>0$ depending only on $M$ such that 
$$
P\left(\sup_{x\in\mathbb{R}}\left|\tilde{F}(x)-F(x)\right|\leq\delta_{1n}\right)\geq1-\gamma_{n},
$$
where $\gamma_{n}= M'(\log K)^2/(K\delta_{1n})  $. 
	\item Let $\Pi=\Pi_{\text{OB}} $  the set of overlapping blocks defined in \eqref{eq: overlapping permutation}. Suppose that for each $\pi\in \Pi_{\text{CSO}} $, there exist a further permutation $\{u^{\pi}_t\}_{t=1}^K=\{S_*(Z^{\pi \circ \tilde{\pi}} )\}_{\tilde{\pi}\in\Pi_{\text{NOB}}} $ such that $\{u^{\pi}_t\}_{t=1}^{K_\pi} $ and $\{u^{\pi}_t\}_{t=K_\pi+2}^{K} $ are stationary and strong mixing with $ \sum_{k=1}^\infty \alpha_{\text{mixing}}(k) \leq M $ for a constant $M$ that does not depend on $\pi$. Then   there exists a constant $M'>0$ depending only on $M$ such that 
$$
P\left(\sup_{x\in\mathbb{R}}\left|\tilde{F}(x)-F(x)\right|\leq\delta_{1n}\right)\geq1-\gamma_{n},
$$
where $\gamma_{n}= M'(\log K)/(\sqrt{K}\delta_{1n})  $. 
\end{enumerate}
\end{lem}

Strong mixing is a mild condition on dependence and is satisfied by many stochastic processes. For example, it is well known that any stationary Markov chains that are Harris recurrent and aperiodic are strong mixing. Many common serially dependent processes such as ARMA with i.i.d. innovations can also be shown to be strong mixing.





\section{Conclusion}
\label{sec: conclusion}
This paper extends the applicability of conformal inference to general settings that allow for time series data. Our results are developed within the general framework of randomization inference. Our method is based on a carefully-designed randomization approach based on groups of permutations, which exhibit a block structure to account for the potential serial dependence in the data. When the data are i.i.d.\ or more generally exchangeable, our method exhibits exact, model-free validity. When the exchangeability condition does not hold, finite-sample performance bounds can still be obtained under weak conditions on the conformity score as long as transformations of the data serve as meaningful approximations for a stationary series. 

\section*{Acknowledgements}
We gratefully acknowledge research support from the National Science Foundation. We are very grateful to three anonymous referees for helpful comments.

\bibliographystyle{apalike}
\bibliography{SC_biblio}

\newpage
\appendix

\section{Proof of Theorem \ref{prop: general exact validity}} The proof essentially follows by
standard arguments, see, e.g.   \cite{romano1990behavior}. We have by \eqref{quant bound}
$$
\sum_{\pi \in \Pi} \mathbf{1}( S(Z^\pi ) > S^{(k)}(Z^\pi) ) \leq \sum_{\pi \in \Pi} \mathbf{1}( S(Z^\pi ) > S^{(k)}(Z)) \leq \alpha n.
$$
Since $\mathbf{1}( S(Z) > S^{(k)}(Z) )$ is equal in law to  $\mathbf{1}( S(Z^\pi) > S^{(k)}(Z^\pi) )$  for any $\pi \in \Pi$ by the exchangeability hypothesis, we have that
$$
\alpha \geq E \sum_{\pi \in \Pi} \mathbf{1}( S(Z^\pi) > S^{(k)}(Z^\pi) )/n = E  \mathbf{1}( S(Z) > S^{(k)}(Z) ) = E \mathbf{1} (\hat p \leq \alpha).
$$

\section{Proof of Theorem \ref{thm: high level new}}

Since the second claim (bounds on the coverage probability) is implied by the first claim, it suffices to show the first claim. Define $$ \hat{F}(x)=\frac{1}{n} \sum_{\pi\in\Pi} \mathbf{1}\{S(Z^\pi)<x \}. $$
The rest of the proof proceeds in two steps. We first bound $\hat{F}(x)-F(x) $ and then derive the desired result.

\textbf{Step 1:} We bound the difference between the $p$-value and the oracle $p$-value, $\hat{F}(S(Z))-F(S_*(Z))$.

Let $\mathcal{M}$ be the event that the conditions (A) and (E) hold. By assumption, 
\begin{equation}
P\left(\mathcal{M}\right)\geq1-\gamma_{1n}-\gamma_{2n}.\label{eq: thm high level eq 0.5}
\end{equation}

Notice that on the event $\mathcal{M}$, 
\begin{align}
\left|\hat{F}(S(Z))-F(S_*(Z))\right| & \leq\left|\hat{F}(S(Z))-F(S(Z))\right|+\left|F(S(Z))-F(S_*(Z))\right|\nonumber \\
 & \overset{\text{ (i)}}{\leq}\sup_{x\in\mathbb{R}}\left|\hat{F}(x)-F(x)\right|+D\left|S(Z)-S_*(Z)\right|\nonumber \\
 & \leq\sup_{x\in\mathbb{R}}\left|\hat{F}(x)-\tilde{F}(x)\right|+\sup_{x\in\mathbb{R}}\left|\tilde{F}(x)-F(x)\right|+D\left|S(Z)-S_*(Z)\right|\nonumber \\
 & \leq\sup_{x\in\mathbb{R}}\left|\hat{F}(x)-\tilde{F}(x)\right|+\delta_{1n}+D\left|S(Z)-S_*(Z)\right|\nonumber \\
 & \leq\sup_{x\in\mathbb{R}}\left|\hat{F}(x)-\tilde{F}(x)\right|+\delta_{1n}+D\delta_{2n},\label{eq: thm high level eq 1}
\end{align}
where (i) holds by the fact that the bounded pdf of $S_*(Z)$ implies
Lipschitz property for $F$. 

Let $A=\left\{ \pi\in\Pi:\ |S(Z^\pi)-S_*(Z^\pi)|\geq\sqrt{\delta_{2n}}\right\} $.
Observe that on the event $\mathcal{M}$, by Chebyshev inequality
\[
|A|\delta_{2n}\leq\sum_{\pi\in\Pi}\left(S(Z^\pi)-S_*(Z^\pi)\right)^{2}\leq n\delta_{2n}^{2}
\]
and thus $|A|/n\leq\delta_{2n}$. Also observe that on the event $\mathcal{M}$,
for any $x\in\mathbb{R}$, 
\begin{align}
 & \left|\hat{F}(x)-\tilde{F}(x)\right| \nonumber \\
 & \leq\frac{1}{n}\sum_{\pi\in A}\left|\mathbf{1}\left\{ S(Z^\pi)< x\right\} -\mathbf{1}\left\{ S_*(Z^\pi)< x \right\} \right|+\frac{1}{n}\sum_{\pi\in(\Pi\backslash A)}\left|\mathbf{1}\left\{ S(Z^\pi)< x\right\} -\mathbf{1}\left\{ S_*(Z^\pi)< x\right\} \right|\nonumber \\
 & \overset{\mathrm{(i)}}{\leq}2\frac{|A|}{n}+\frac{1}{n}\sum_{\pi\in(\Pi\backslash A)}\mathbf{1}\left\{ \left|S_*(Z^\pi)-x\right|\leq\sqrt{\delta_{2n}}\right\}  \leq2\frac{|A|}{n}+\frac{1}{n}\sum_{\pi\in\Pi}\mathbf{1}\left\{ \left|S_*(Z^\pi)-x\right|\leq\sqrt{\delta_{2n}}\right\} \nonumber \\
 & \leq2\frac{|A|}{n}+P\left(\left|S_*(Z)-x\right|\leq\sqrt{\delta_{2n}}\right)+\sup_{z\in\mathbb{R}}\left|\frac{1}{n}\sum_{\pi\in\Pi}\mathbf{1}\left\{ \left|S_*(Z^\pi)-z\right|\leq\sqrt{\delta_{2n}}\right\} -P\left(\left|S_*(Z)-z\right|\leq\sqrt{\delta_{2n}}\right)\right| \nonumber \\
 & =2\frac{|A|}{n}+P\left(\left|S_*(Z)-x\right|\leq\sqrt{\delta_{2n}}\right) \nonumber \\
 & \qquad+\sup_{x\in\mathbb{R}}\left|\left[\tilde{F}\left(z+\sqrt{\delta_{2n}}\right)-\tilde{F}\left(z-\sqrt{\delta_{2n}}\right)\right]-\left[F\left(z+\sqrt{\delta_{2n}}\right)-F\left(z-\sqrt{\delta_{2n}}\right)\right]\right| \nonumber \\
 & \leq2\frac{|A|}{n}+P\left(\left|S_*(Z)-x\right|\leq\sqrt{\delta_{2n}}\right)+2\sup_{z\in\mathbb{R}}\left|\tilde{F}(z)-F\left(z\right)\right|\nonumber  \\
 & \overset{\mathrm{(ii)}}{\leq}2\frac{|A|}{n}+2D\sqrt{\delta_{2n}}+2\delta_{1n} \overset{\mathrm{(iii)}}{\leq }2\delta_{1n}+ 2 \delta_{2n}+ 2 D\sqrt{\delta_{2n}}, \label{eq: bound ecdf error}
\end{align}
where (i) follows by the boundedness of indicator functions and
the elementary inequality of $|\mathbf{1}\{S(Z^\pi)< x\}-\mathbf{1}\{S_*(Z^\pi)<x\}|\leq\mathbf{1}\{|S_*(Z^\pi)-x|\leq|S(Z^\pi)-S_*(Z^\pi)|\}$,
(ii) follows by the bounded pdf of $S_*(Z)$ and (iii) follows by $|A|/n\leq\delta_{2n}$.
Since the above display holds for each $x\in\mathbb{R}$, it follows
that on the event $\mathcal{M}$, 
\begin{equation}
\sup_{x\in\mathbb{R}}\left|\hat{F}(x)-\tilde{F}(x)\right|\leq
2\delta_{1n}+ 2 \delta_{2n}+ 2D\sqrt{\delta_{2n}}.\label{eq: thm high level eq 2}
\end{equation}

We combine (\ref{eq: thm high level eq 1}) and (\ref{eq: thm high level eq 2})
and obtain that on the event $\mathcal{M}$, 
\begin{equation}
\left|\hat{F}(S(Z))-F(S_*(Z))\right|\leq
3\delta_{1n}+ 2 \delta_{2n}+ D(\delta_{2n}+2 \sqrt{\delta_{2n}}).\label{eq: thm high level eq 3}
\end{equation}

\textbf{Step 2:} Here we derive the desired result. Notice that 
\begin{align*}
& \left|P\left(1-\hat{F}(S(Z))\leq\alpha\right)-\alpha\right|\\
 &  =\left|E\left(\mathbf{1}\left\{ 1-\hat{F}(S(Z))\leq\alpha\right\} -\mathbf{1}\left\{ 1-F(S_*(Z))\leq\alpha\right\} \right)\right|\\
 & \leq E\left|\mathbf{1}\left\{ 1-\hat{F}(S(Z))\leq\alpha\right\} -\mathbf{1}\left\{ 1-F(S_*(Z))\leq\alpha\right\} \right|\\
 & \overset{\mathrm{(i)}}{\leq}P\left(\left|F(S_*(Z))-1+\alpha\right|\leq\left|\hat{F}(S(Z))-F(S_*(Z))\right|\right)\\
 & \leq P\left(\left|F(S_*(Z))-1+\alpha\right|\leq\left|\hat{F}(S(Z))-F(S_*(Z))\right|\ \text{ and}\ \mathcal{M}\right)+P(\mathcal{M}^{c})\\
 & \overset{\mathrm{(ii)}}{\leq}P\left(\left|F(S_*(Z))-1+\alpha\right|\leq
  3\delta_{1n}+ 2 \delta_{2n}+ D(\delta_{2n}+ 2\sqrt{\delta_{2n}})\right)+P\left(\mathcal{M}^{c}\right)\\
 & \overset{\mathrm{(iii)}}{\leq}
6\delta_{1n}+ 4 \delta_{2n}+ 2 D(\delta_{2n}+ 2\sqrt{\delta_{2n}}) +\gamma_{1n}+\gamma_{2n},
\end{align*}
where (i) follows by the elementary inequality $|\mathbf{1}\{1-\hat{F}(S(Z))\leq\alpha\}-\mathbf{1}\{1-F(S_*(Z))\leq\alpha\}|\leq\mathbf{1}\{|F(S_*(Z))-1+\alpha|\leq|\hat{F}(S(Z))-F(S_*(Z))|\}$,
(ii) follows by (\ref{eq: thm high level eq 3}), (iii) follows
by the fact that $F(S_*(Z))$ has the uniform distribution on $(0,1)$
and hence has pdf equal to 1, and  by (\ref{eq: thm high level eq 0.5}).
The proof is complete.

\section{Proof of Lemma \ref{lem: primitive cond for ergodicity}}
\textbf{Proof of the first claim.} By assumption, 
\[
\tilde{F}(x)-F(x)=\frac{1}{K}\sum_{t=1}^{K}\left(\mathbf{1}\{u_{t}<x\}-F(x)\right).
\]

Applying Proposition 7.1 of  \cite{rio2017asymptotic}, we have that 
\[
E\left(\sup_{x\in\mathbb{R}}\left|\tilde{F}(x)-F(x)\right|^{2}\right)\leq\frac{1+4M}{K}\left(3+\frac{\log K}{2\log2}\right)^{2}.
\]

Therefore, the first result follows by Markov's inequality 
\begin{align*}
P\left(\sup_{x\in\mathbb{R}}\left|\tilde{F}(x)-F(x)\right|>\delta_{1n}\right) & \leq\frac{E\left(\sup_{x\in\mathbb{R}}\left|\tilde{F}(x)-F(x)\right|^{2}\right)}{\delta_{1n}^{2}}\\
 & \le\frac{1+4M}{K\delta_{1n}^{2}}\left(3+\frac{\log K}{2\log2}\right)^{2}.
\end{align*}

\textbf{Proof of the second claim.} For any $\pi\in\Pi_{\text{CSO}}$, define 
\[
G_{\pi}(x)=\frac{1}{K}\sum_{\tilde{\pi}\in\Pi_{\text{NOB}}}\left(\mathbf{1}\{S_{*}(Z^{\pi\circ \tilde{\pi}})<x\}-F(x)\right).
\]

Notice that 
\[
\tilde{F}(x)-F(x)=\frac{1}{T}\sum_{\pi\in\Pi_{\text{CSO}}}G_{\pi}(x).
\]

It follows that 
\begin{equation}
E\sup_{x\in\mathbb{R}}\left|\tilde{F}(x)-F(x)\right|\leq\frac{1}{T}\sum_{\pi\in\Pi_{\text{CSO}}}E\sup_{x\in\mathbb{R}}\left|G_{\pi}(x)\right|.\label{eq: ergodicity eq 2}
\end{equation}

We now bound $E\sup_{x\in\mathbb{R}}\left|G_{\pi}(x)\right|$. For
a fixed $\pi\in\Pi_{\text{CSO}}$, we have
$$G_{\pi}(x)=\frac{1}{K}\sum_{t=1}^{K}\left(\mathbf{1}\{u_{t}^{\pi}<x\}-F(x)\right).
$$
We can further decompose $$
G_{\pi}(x)=\frac{1}{K}\left[K_{\pi}G_{\pi}^{(1)}(x)+(K-K_{\pi}-1)G_{\pi}^{(2)}(x)+\left(\mathbf{1}\{u_{K_{\pi}+1}^{\pi}<x\}-F(x)\right)\right],$$
where $$G_{\pi}^{(1)}(x)=K_{\pi}^{-1}\sum_{t=1}^{K_{\pi}}\left(\mathbf{1}\{u_{t}^{\pi}<x\}-F(x)\right),$$
$$G_{\pi}^{(2)}(x)=(K-K_{\pi}-1)^{-1}\sum_{t=K_{\pi}+2}^{K}\left(\mathbf{1}\{u_{t}^{\pi}<x\}-F(x)\right).$$
By the same argument as in part 1, we can show that $$
E\left(\sup_{x\in\mathbb{R}}\left|G_{\pi}^{(1)}(x)\right|^{2}\right)\leq\frac{1+4M}{K_{\pi}}\left(3+\frac{\log K_{\pi}}{2\log2}\right)^{2},
$$$$
E\left(\sup_{x\in\mathbb{R}}\left|G_{\pi}^{(2)}(x)\right|^{2}\right)\leq\frac{1+4M}{K-K_{\pi}-1}\left(3+\frac{\log(K-K_{\pi}-1)}{2\log2}\right)^{2}.
$$

Let $z\mapsto f(z)$ be defined by 
\[
f(z)=\frac{1+4M}{z}\left(3+\frac{\log(z)}{2\log2}\right)^{2}.
\]

It is not difficult to verify that $d^{2}f(z)/dz^{2}<0$ for $z\geq1$.
Therefore, $f(z)$ is concave on $[1,K-1]$. Therefore, 
\[
f(K_{\pi})+f(K-K_{\pi}-1)\leq2f((K-1)/2)=\frac{4+16M}{K-1}\left(3+\frac{\log((K-1)/2)}{2\log2}\right)^{2}.
\]

It follows that 
\[
E\left(\sup_{x\in\mathbb{R}}\left|G_{\pi}^{(1)}(x)\right|^{2}\right)+E\left(\sup_{x\in\mathbb{R}}\left|G_{\pi}^{(2)}(x)\right|^{2}\right)\leq\frac{4+16M}{K-1}\left(3+\frac{\log((K-1)/2)}{2\log2}\right)^{2}.
\]

Therefore, 
\begin{align*}
E\sup_{x\in\mathbb{R}}\left|G_{\pi}(x)\right|^{2} & =E\sup_{x\in\mathbb{R}}\left|\frac{1}{K}\left[K_{\pi}G_{\pi}^{(1)}(x)+(K-K_{\pi}-1)G_{\pi}^{(2)}(x)+\left(\mathbf{1}\{u_{K_{\pi}+1}^{\pi}<x\}-F(x)\right)\right]\right|^{2}\\
 & \leq E\left(\frac{1}{K}\left[K_{\pi}\sup_{x\in\mathbb{R}}\left|G_{\pi}^{(1)}(x)\right|+(K-K_{\pi}-1)\sup_{x\in\mathbb{R}}\left|G_{\pi}^{(2)}(x)\right|+2\right]\right)^{2}\\
 & \overset{\text{(i)}}{\leq}4\frac{K_{\pi}^{2}}{K^{2}}E\sup_{x\in\mathbb{R}}\left|G_{\pi}^{(1)}(x)\right|^{2}+4\frac{(K-K_{\pi}-1)^{2}}{K^{2}}E\sup_{x\in\mathbb{R}}\left|G_{\pi}^{(2)}(x)\right|^{2}+\frac{8}{K^{2}}\\
 & \leq4\left(E\sup_{x\in\mathbb{R}}\left|G_{\pi}^{(1)}(x)\right|^{2}+E\sup_{x\in\mathbb{R}}\left|G_{\pi}^{(2)}(x)\right|^{2}\right)+\frac{8}{K^{2}}\\
 & \leq\frac{16+64M}{K-1}\left(3+\frac{\log((K-1)/2)}{2\log2}\right)^{2}+\frac{8}{K^{2}},
\end{align*}
where (i) follows by the elementary inequality $(a+b+c)^{2}\leq4a^{2}+4b^{2}+2c^{2}$. 

Notice that the above bound does not depend on $\pi$. In light of
(\ref{eq: ergodicity eq 2}), it follows that 
\[
E\sup_{x\in\mathbb{R}}\left|\tilde{F}(x)-F(x)\right|\leq\sqrt{\frac{16+64M}{K-1}\left(3+\frac{\log((K-1)/2)}{2\log2}\right)^{2}+\frac{8}{K^{2}}}.
\]

The second claim of the lemma follows by Markov's inequality.

\section{Empirical Study}
Here we provide some simulation evidence on the empirical properties of the conformal prediction intervals. We consider a penalized regression setting as in Section \ref{sec: penalized regression} and similar to \citet{lei2017distributionfree}. The data are generated as
\begin{equation*}
Y_t=X_t'\beta+\varepsilon_t, \quad 1\le t \le T,
\end{equation*}
where the features $X_t$ are distributed as $N(0,I_p)$ and independent over time. To induce serial dependence, we generate the error $\varepsilon_t$ based on an AR(1) model with parameter $\rho$:
\[
\varepsilon_t = \rho \varepsilon_{t-1}+\xi_t, ~~ \xi_t\overset{iid}\sim N(0,1-\rho^2), ~~ \text{given }\varepsilon_0=0.
\]
We set $\beta  \propto (1,1,1,1,1,0,\dots,0)'$ with $\|\beta\|_2=2$. The number of features is $p=100$. For simplicity, we let $T_1=1$ and choose the block size to be $b=1$ such that the number of blocks is $K=T$. The coefficients $\beta$ are estimated using LASSO as implemented in the \textrm{R} package \texttt{hdm} \citep{hdm}. We choose a residual based test statistic as in \citet{lei2017distributionfree}: 
\begin{eqnarray*}
S(Z)=|Y_T-X_T'\hat\beta (Z)|,
\end{eqnarray*}
where $\hat\beta(Z)$ is the estimate of $\beta$ based on the augmented data $Z$.
Confidence sets $\mathcal{C}_{1-\alpha}$ are computed based on Algorithm \ref{proc: generalized CP}. We choose $\Pi$ to be the set of non-overlapping block permutations $\Pi_{NOB}$. The number of grid points is $H=100$.

\begin{figure}[h]
\begin{center}
\includegraphics[width=0.45\textwidth]{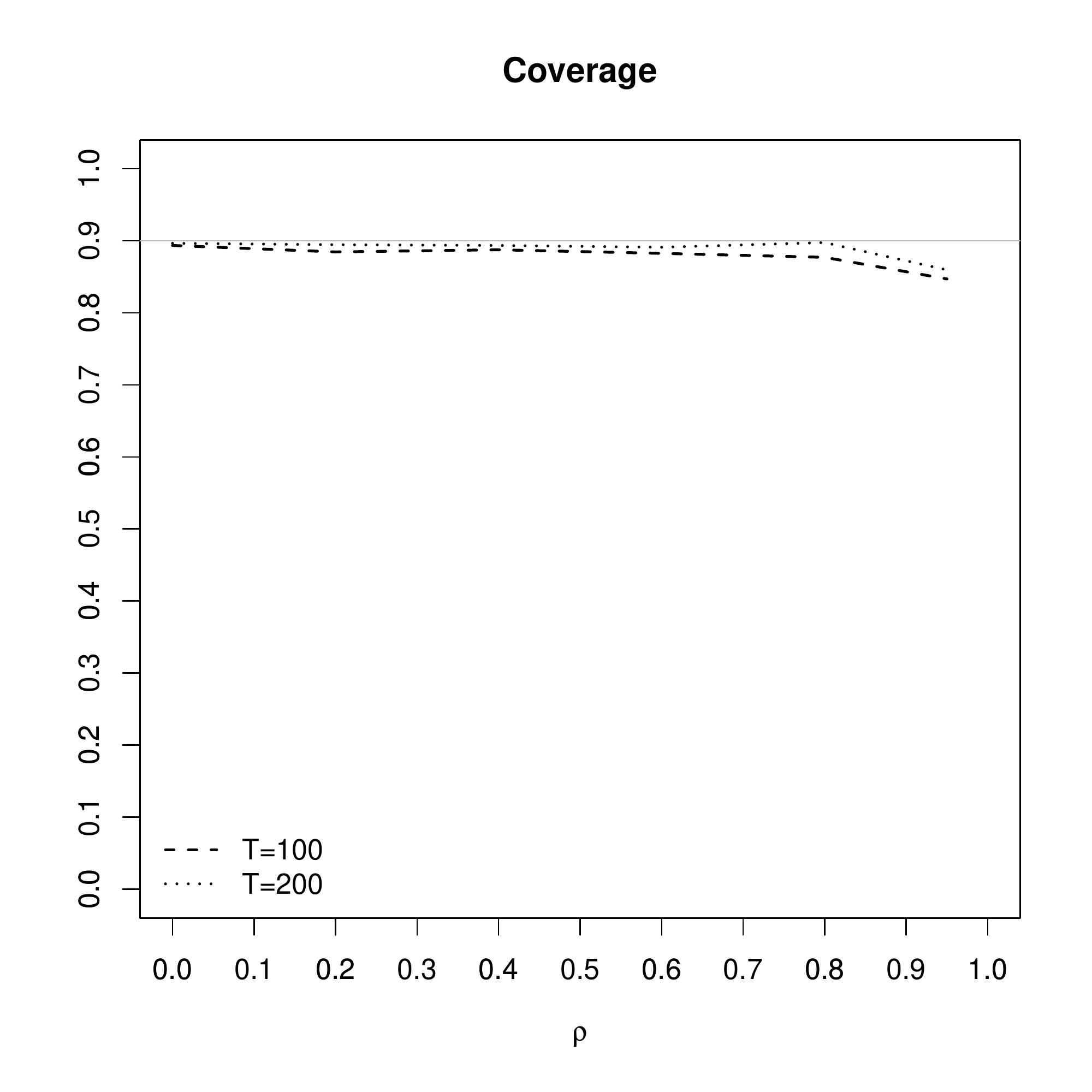}
\includegraphics[width=0.45\textwidth]{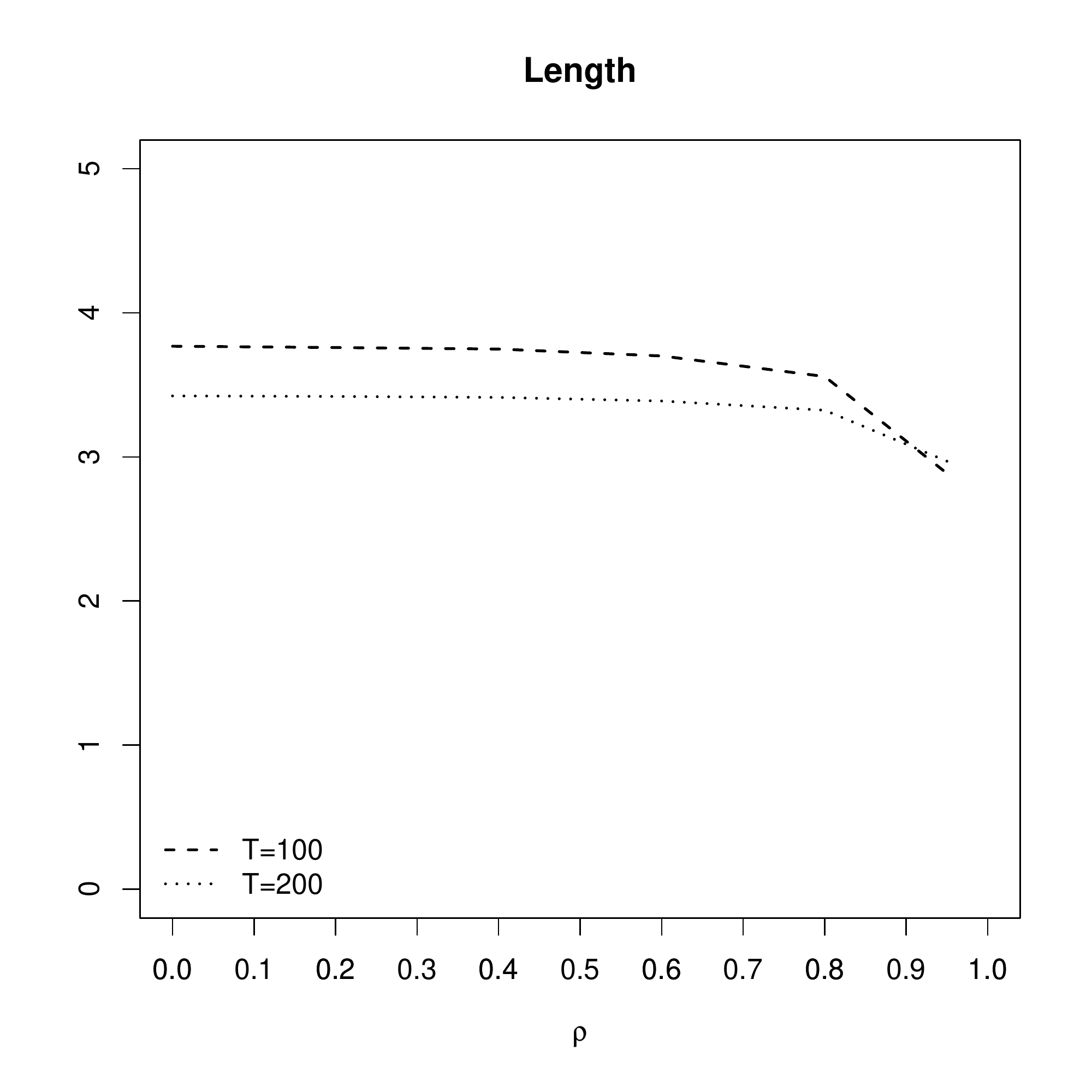}
\end{center}
\caption{The number of simulations is 2000.}
\label{fig:coverage}
\end{figure}

Figure \ref{fig:coverage} displays the empirical coverage rates and the average length of the confidence intervals for $T\in \{100,200\}$ and different values of $\rho$ between $0$ and $0.95$. For the case where the data are independent over time ($\rho=0$), Theorem \ref{prop: general exact validity} asserts that our procedure enjoys exact finite sample validity, which is confirmed by the simulation results. When the data exhibit serial correlation ($\rho>0$), Theorem \ref{thm: high level new} asserts that our method achieves approximate validity. The simulation results show that, for a wide range of values of $\rho$, the empirical coverage rates are very close to the nominal coverage rate of $1-\alpha=0.9$. Only for very high values of $\rho$, the confidence intervals exhibit some undercoverage. The average length of the confidence intervals is relatively constant up to around $\rho=0.6$ and decreasing at an increasing rate for higher values of $\rho$. Moreover, the average length is decreasing in the sample size $T$, except for very high values of $\rho$.

Overall, the simulation results demonstrate that our procedure exhibits favorable finite sample properties in settings where the data exhibit times series dependence.

\end{document}